\documentclass[runningheads]{llncs}

\usepackage[T1]{fontenc}
\def\doi#1{\href{https://doi.org/\detokenize{#1}}{\url{https://doi.org/\detokenize{#1}}}}
\usepackage{graphicx}
\usepackage{xcolor}
\usepackage{amsmath}
\usepackage{amsfonts}
\usepackage{hyperref}

\usepackage{cleveref}
\usepackage{colortbl}
\usepackage{multirow}
\usepackage{multicol}
\usepackage{makecell}
\usepackage{booktabs}
\usepackage{marvosym}
\usepackage{subcaption}
\usepackage{url}
\definecolor{mygray}{gray}{0.92}
\definecolor{mygray2}{gray}{0.85}

\usepackage{listings}
\begin{document}
\title{SLPD: Slide-level Prototypical Distillation \\ for WSIs}
\titlerunning{SLPD}

\author{Zhimiao Yu\inst{\star} 
\and
Tiancheng Lin\inst{\star} 
\and
Yi Xu\textsuperscript{(\Letter)}} 

\institute{MoE Key Lab of Artificial Intelligence, AI Institute, \\ Shanghai Jiao Tong University, Shanghai, China
 \\ \email{xuyi@sjtu.edu.cn}}
\maketitle              
\begin{abstract}
Improving the feature representation ability is the foundation of many whole slide pathological image (WSIs) tasks. Recent works have achieved great success in pathological-specific self-supervised learning (SSL). However, most of them only focus on learning patch-level representations, thus there is still a gap between pretext and slide-level  downstream tasks, $e.g.$, subtyping, grading and staging. Aiming towards slide-level representations, we propose Slide-Level Prototypical Distillation (SLPD) to explore intra- and inter-slide semantic structures  for context modeling on WSIs. Specifically, we iteratively perform intra-slide clustering for the regions (4096$\times$4096 patches) within each WSI to yield the  prototypes and  encourage the region representations to be closer to the assigned prototypes.
By representing each slide with its prototypes, we further select similar slides by the set distance of prototypes and assign the regions by cross-slide prototypes for distillation.
SLPD achieves state-of-the-art results on multiple slide-level benchmarks and demonstrates that representation learning of semantic structures of slides can make a suitable proxy task for WSI analysis. Code will be available at \href{https://github.com/Carboxy/SLPD}{https://github.com/Carboxy/SLPD}.

\keywords{Computational pathology \and Whole slide images(WSIs)  \and Self-supervised learning.}
\end{abstract}
\section{Introduction}
\footnote{$\star$: Equal contribution.}
In computational histopathology, visual representation extraction is a fundamental problem~\cite{histofeature}, serving as a cornerstone of the (downstream) task-specific learning on whole slide pathological images (WSIs). 
Our community has witnessed the progress of the \emph{de facto} representation learning paradigm from the supervised ImageNet pre-training 
to self-supervised learning (SSL)~\cite{wu2018unsupervised,he2020momentum}. 
Numerous pathological applications benefit from SSL, including classification of glioma~\cite{chen2019self}, breast carcinoma~\cite{abbasi2021molecular}, and non-small-cell lung carcinoma~\cite{li2023self}, mutation prediction~\cite{saldanha2022self}, microsatellite instability prediction~\cite{saillard2021self}, 
and survival prediction from WSIs~\cite{huang2021integration,abbet2020divide}. 
Among them, pioneering works~\cite{li2021dual,lu2020,dehaene2020} 
directly apply the SSL algorithms  developed for natural images ($e.g.$, SimCLR~\cite{simclr}, CPC~\cite{cpc} and MoCo~\cite{mocov2}) to WSI analysis tasks, and the improved performance proves the effectiveness of SSL.
However, WSI is quite different from natural images in that it exhibits a hierarchical structure with giga-pixel resolution.
Following works turn to \textit{designing pathological-specific tasks} to explore the inherent characteristics of WSIs for representation learning, $e.g.$, resolution-aware tasks~\cite{selfpath,BSP,xie2020} and color-aware tasks~\cite{abbet2020divide,csco}.
Since the pretext tasks encourage to mine the pathologically relevant patterns, the learned representations are expected to be more suitable for WSI analysis. 
Nevertheless, these works only consider learning the representations at the patch level, $i.e$, the cellular organization, but neglecting macro-scale morphological features, $e.g.$, tissue phenotypes and intra-tumoral heterogeneity. As a result, there is still a gap between the pre-trained representations and  downstream tasks, as the latter is mainly at the slide level, $e.g.$, subtyping, grading and staging. 

More recently, some works propose to close the gap via \textit{directly learning slide-level representations in pre-training}.
For instance, HIPT~\cite{chen2022scaling}, a milestone work, introduces hierarchical pre-training (DINO~\cite{dino}) for the patch-level (256$\times$256) and region-level (4096$\times$4096) in a two-stage manner, achieving superior performance on slide-level tasks.
SS-CAMIL~\cite{ss-camil} uses EfficientNet-B0 for image compression in the first stage and then derives multi-task learning on the compressed WSIs, which assumes the primary site information, $e.g.$, the organ type, is always available and can be used as pseudo labels.
SS-MIL~\cite{ss-mil} also proposes a two-stage pre-training framework for WSIs using contrastive learning (SimCLR~\cite{simclr}), where the differently subsampled bags~\footnote{By formulating WSI tasks as a multi-instance learning problem, the WSI is treated as a bag with corresponding patches as instances.} from the same WSI are positive pairs in the second stage.
A similar idea can be found in Giga-SSL~\cite{giga-ssl} with delicate patch- and WSI-level augmentations.
The aforementioned methods share the same two-stage pre-training paradigm, $i.e.$, patch-to-region/slide. Thus broader context information is preserved to close the gap between pretext and downstream tasks. 
However, they are essentially \textit{instance discrimination} where only the self-invariance of region/slide is considered, leaving the \textit{intra- and inter-slide semantic structures} unexplored.

In this paper, we propose to encode the intra- and inter-slide semantic structures by modeling the mutual-region/slide relations, which is called SLPD: Slide-Level Prototypical Distillation for WSIs.
Specifically, we perform the slide-level clustering for the 4096$\times$4096 regions within each WSI to yield the prototypes, which characterize the medically representative patterns of the tumor ($e.g.$, morphological phenotypes). In order to learn this intra-slide semantic structure, 
we encourage the region representations to be closer to the assigned prototypes.
By representing each slide with its prototypes, we further select semantically similar slides by the set-to-set distance of prototypes. Then, we learn the inter-slide semantic structure by building correspondences between region representations and cross-slide prototypes. 
We conduct experiments on two benchmarks, NSCLC subtyping and BRCA subtyping.
SLPD achieves state-of-the-art results on multiple slide-level tasks, demonstrating that representation learning of semantic structures of slides can make a suitable proxy task for WSI analysis. We also perform extensive ablation studies to verify the effectiveness of crucial model components. 
\begin{figure}[t]
\centering
\includegraphics[scale=0.41]{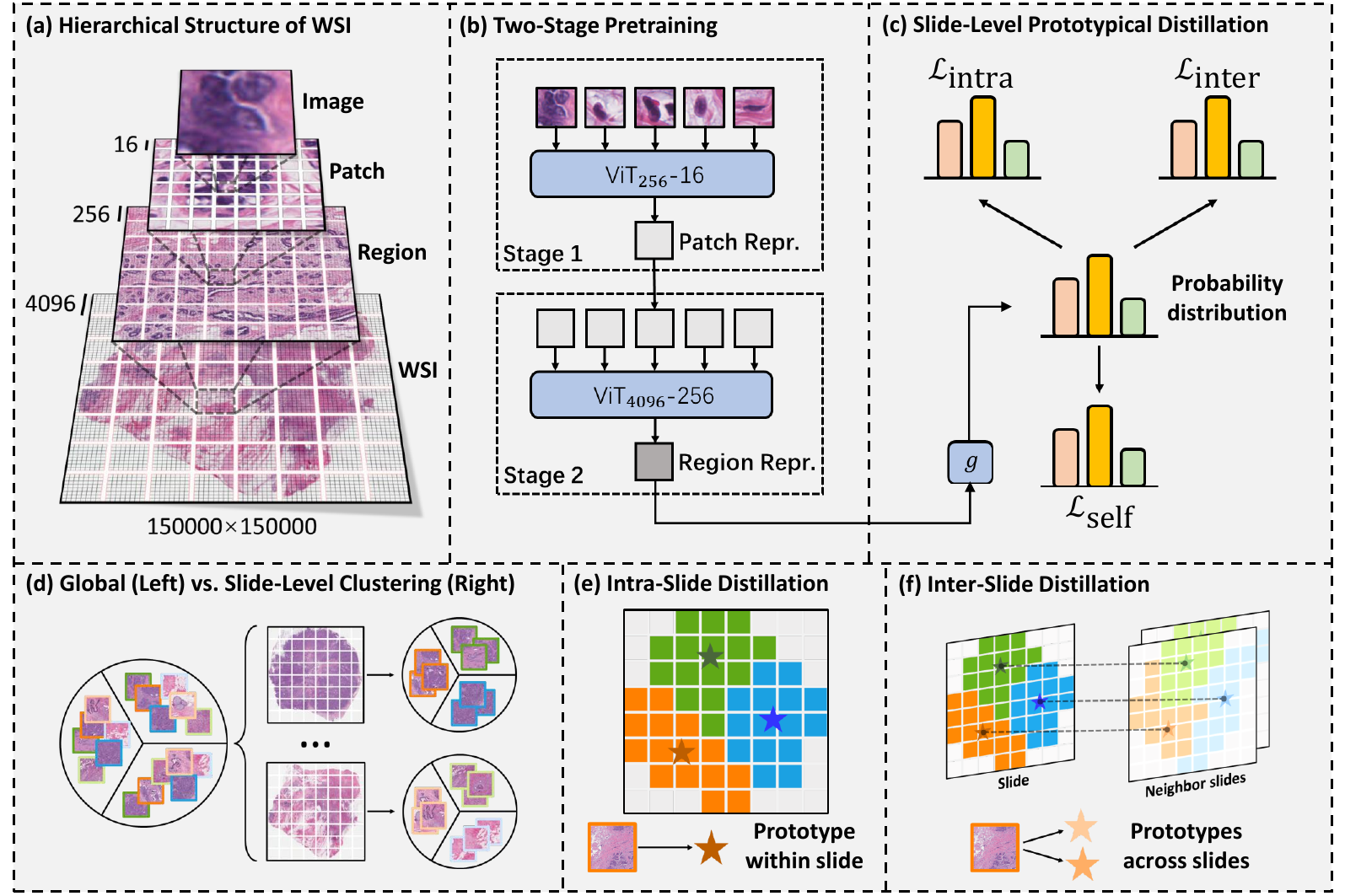}
\caption{(a) A WSI possesses the hierarchical structure of WSI-region-patch-image, from coarse to fine. (b) Two-stage pre-training paradigm successively performs the image-to-patch and patch-to-region aggregations. (c-e) The proposed SLPD. SLPD explores the semantic structure by slide-level clustering. Besides self-distillation, region representations are associated with the prototypes within and across slides to comprehensively understand WSIs.}
\label{fig:main}
\end{figure}
\section{Method}
\subsection{Overview}

As shown in Fig~\ref{fig:main}(a), a WSI exhibits hierarchical structure at varying resolutions under $20\times$ magnification: 1) the $4096\times 4096$ regions describing macro-scale organizations of cells, 2) the $256\times 256$ patches capturing local clusters of cells, 3) and the $16\times 16$ images characterizing the fine-grained features at the cell-level. Given $N$ unlabeled WSIs $\{w_1,w_2,\cdots, w_N \}$, consisting of numerous regions $\{\{x_n^l \}_{l=1}^{L_n}\}_{n=1}^N$, where $L_n$ denotes the number of regions of WSI $w_n$, we aim to learn a powerful encoder that maps each $x_n^l$ to an embedding $z_n^l \in \mathbb{R}^D$.
SLPD is built upon the two-stage pre-training paradigm proposed by HIPT, which will be described in Sec.~2.2. Fig~\ref{fig:main}(c-d) illustrates the pipeline of SLPD.
We characterize the semantic structure of slides in Sec.~2.3, which is leveraged to establish the relationship within and across slides, leading to the proposed intra- and inter-slide distillation in Sec.~2.4 and Sec.~2.5.

\subsection{Preliminaries}
\label{sec:pre}
We revisit Hierarchical Image Pyramid Transformer (HIPT)~\cite{chen2022scaling}, a cutting-edge method for learning representations of WSIs via self-supervised vision transformers. As shown in Fig.~\ref{fig:main}(b), HIPT proposes a two-stage pre-training paradigm considering the hierarchical structure of WSIs. In stage one, a patch-level vision transformer, denoted as ViT$_{256}$-16, aggregates non-overlapping $16\times16$ images within $256\times256$ patches to form patch-level representations. In stage two, the pre-trained ViT$_{256}$-16 is freezed and leveraged to tokenize the patches within $4096\times4096$ regions. Then a region-level vision transformer ViT$_{4096}$-256 aggregates these tokens to obtain region-level representations. With this hierarchical aggregation strategy, a WSI can be represented as a bag of region-level representations, which are then aggregated with another  vision transformer, ViT$_{\text{WSI}}$-4096, to perform slide-level prediction tasks.

HIPT leverages DINO~\cite{dino} to pre-train ViT$_{256}$-16 and ViT$_{4096}$-256, respectively. The learning objective of DINO is self-distillation. Taking stage two as an example, DINO distills the knowledge from teacher to student by minimizing the cross-entropy between the probability distributions of two views at region-level:


\begin{equation}
    \mathcal{L}_{\text{self}} = \mathbb{E}_{x\sim p_d} H(g_t(\hat{z}), g_s(z)),
\end{equation}
where $H(a,b)=-a\log b$, and $p_d$ is the data distribution that all regions are drawn from. The teacher and the student share the same architecture consisting of an encoder ($e.g.$, ViT) and a projection head $g_t/g_s$. $\hat{z}$ and $z$ are the embeddings of two views at region-level yielded by the encoder. 
The parameters of the student are exponentially moving averaged to the parameters of the teacher.




\subsection{Slide-Level Clustering}
Many histopathologic features have been established based on the morphologic phenotypes of the tumor, such as tumor invasion, anaplasia, necrosis and mitoses, which are then used for cancer diagnosis, prognosis and the estimation of response-to-treatment in patients~\cite{amin2017eighth,chen2022pan}. To obtain meaningful representations of slides, we aim to explore and maintain such histopathologic features in the latent space. Clustering can reveal the representative patterns in the data and has achieved success in the area of unsupervised representation learning~\cite{caron2018deep,caron2020unsupervised,li2020prototypical,li2021contrastive}. To characterize the histopathologic features underlying the slides, a straightforward practice is the global clustering, $i.e.$,
clustering the region embeddings from all the WSIs, as shown in the left of Fig.~\ref{fig:main}(d). 
However, the obtained clustering centers, $i.e.$, the prototypes, are inclined to represent the visual bias related to staining or scanning procedure rather than medically relevant features~\cite{sharma2021cluster}.
Meanwhile, this clustering strategy ignores the hierarchical structure ``region$\rightarrow$WSI$\rightarrow$whole dataset'' underlying the data, where the ID of the WSI can be served as an extra learning signal. 
Therefore, we first consider the slide-level clustering that clusters the embeddings within each WSI, which is shown in the right of Fig.~\ref{fig:main}(d). Specifically, we conduct $k$-means algorithm before the start of each epoch over $L_n$ region embeddings $\{z_n^l \}_{l=1}^{L_n}$ of $w_n$ to obtain $M$ prototypes $\{c_n^m \in \mathbb{R}^D\}_{m=1}^M$.
Similar operations are applied across other slides, and then we acquire $N$ groups of prototypes $\{\{c_n^m\}_{m=1}^M\}_{n=1}^N$. Each group of prototypes is expected to encode the semantic structure ($e.g.$, the combination of histopathologic features) of the WSI.

\subsection{Intra-Slide Distillation}
The self-distillation utilized by HIPT in stage two encourages the correspondence between two views of a region
at the  macro-scale because the organizations of cells share mutual information spatially. However, the self-distillation, which solely mines the spatial correspondences inside the $4096\times 4096$ region, cannot comprehensively understand the histopathologic consistency at the slide-level. 
In order to achieve better representations, the histopathologic connections between the WSI and its regions should be modeled and learned, which is called intra-slide correspondences.
With the proposed slide-level clustering in Sec.~{2.3}, a slide can be abstracted by a group of prototypes, which capture the semantic structure 
of the WSI. As shown in Fig.~\ref{fig:main}(e), we assume that the representation $z$ and its assigned prototype $c$ also share mutual information and encourage $z$ to be closer to $c$ with the intra-slide distillation:
\begin{equation}
\label{eq:intra}
    \mathcal{L}_{\text{intra}} = \mathbb{E}_{x\sim p_d}H\left(g_t(c), g_s(z)\right),
\end{equation}
We omit super-/sub-scripts of $z$ for brevity. Through Eq.~\ref{eq:intra}, we can leverage more intra-slide correspondences to guide the learning process. 
For further understanding, a prototype can be viewed as an augmented representation aggregating the slide-level information. Thus this distillation objective is encoding such information into the corresponding region embedding, which makes the learning process semantic structure-aware at the slide-level.




\subsection{Inter-Slide Distillation}
Tumors of different patients can exhibit morphological similarities in some respects~\cite{jass2004hnpcc,levy2020spatial}, so the correspondences across slides should be characterized during learning. Previous self-supervised learning methods applied to histopathologic images only capture such correspondences with positive pairs at the patch-level~\cite{li2021sslp,li2021dual}, which overlooks the semantic structure of the WSI. We rethink this problem from the perspective how to measure the similarity between two slides accurately. 
Due to the heterogeneity of the slides, comparing them with the local crops or the averaged global features are both susceptible to being one-sided. To address this, we bridge the slides with their semantic structures and 
define the semantic similarity between two slides $w_i$ and $w_j$ through an optimal bipartite matching between two sets of prototypes:
\begin{equation}
\label{eq:match}
    D(w_i,w_j)=\max \{\frac{1}{M}\sum _{m=1}^M \cos (c_i^m, c_j^{\sigma (m)}) ~|~\sigma \in \mathfrak{S}_M \},~ D(w_i,w_j)\in [-1,1],
\end{equation}
where $\cos(\cdot, \cdot)$ measures the cosine similarity between two vectors, and $\mathfrak{S}_M$ enumerates the permutations of $M$ elements. The optimal permutation $\sigma ^*$ can be computed efficiently with the Hungarian algorithm~\cite{kuhn1955hungarian}. 
With the proposed set-to-set distance, we can model the inter-slide correspondences conveniently and accurately. Specifically, for a region embedding $z$ belonging to the slide $w$ and assigned to the prototype $c$, we first search the top-$K$ nearest neighbors of $w$ in the dataset based on the semantic similarity, denoted as $\{\hat{w}_k \}_{k=1}^K$. Second, we also obtain the matched prototype pairs $\{(c, \hat{c}_k)\}_{k=1}^K$ determined by the optimal permutation, where $\hat{c}_k$ is the prototype of $\hat{w}_k$. Finally, we encourage $z$ to be closer to $\hat{c}_k$ with the inter-slide distillation:
\begin{equation}
\label{eq:inter}
    \mathcal{L}_{\text{inter}} = \mathbb{E}_{x\sim p_d}[\frac{1}{K}\sum _{k=1}^K H\left(g_t(\hat{c}_k), g_s(z)\right)].
\end{equation}
The inter-slide distillation can encode the sldie-level information complementary to that of intra-slide distillation into the region embeddings.

The overall learning objective of the proposed SLPD is defined as:
\begin{equation}
\mathcal{L}_{\text{total}} = \mathcal{L}_{\text{self}} + \alpha_1\mathcal{L}_{\text{intra}} +\alpha_2 \mathcal{L}_{\text{inter}},
\end{equation}
where the loss scale is simply set to $\alpha_1=\alpha_2=1$. We believe the performance can be further improved by tuning this.






\section{Experimental Results}
\noindent \textbf{Datasets.}
We conduct experiments on two public WSI datasets~\footnote{The data is released under a CC-BY-NC 4.0 international license.}. \textit{{TCGA-NSCLC}} dataset includes two subtypes in lung cancer, Lung Squamous Cell Carcinoma and Lung Adenocarcinoma, with a total of 1,054 WSIs. \textit{{TCGA-BRCA}} dataset includes two subtypes in breast cancer, Invasive Ductal and Invasive Lobular Carcinoma, with a total of 1,134 WSIs.

\noindent\par
\noindent \textbf{Pre-training.} 
We extract 62,852 and 60,153 regions at $20\times$ magnification from TCGA-NSCLC and TCGA-BRCA for pre-training ViT$_{4096}$-256 in stage two. We leverage the pre-trained ViT$_{256}$-16 in stage one provided by HIPT to tokenize the patches within each region.
Following the official code of HIPT, ViT$_{4096}$-256 is optimized for 100 epochs with optimizer of AdamW, base learning rate of 5e-4 and batch size of 256 on 4 GTX3090 GPUs.

\noindent \textbf{Fine-tuning.}
We use the pre-trained ViT$_{256}$-16 and ViT$_{4096}$-256 to extract embeddings at the \textit{{patch-level}} ($256\times 256$) and the \textit{{region-level}} ($4096\times 4096$) for downstream tasks. With the pre-extracted embeddings,
we fine-tune three aggregators ($i.e.$, MIL~\cite{lu2021data}, DS-MIL~\cite{li2021dual} and ViT$_{\text{WSI}}$-4096~\cite{chen2022scaling}) for 20 epochs and follow other settings in the official code of HIPT.

\noindent \textbf{Evaluation metrics.}
We adopt the 10-fold cross validated Accuracy (Acc.) and area under the curve (AUC) to evaluate the weakly-supervised classification performance. The data splitting scheme is kept consistent with HIPT.

\begin{table}[t]
	\centering
		\caption{Slide-level classification. ``Mean'' leverages the averaged pre-extracted embeddings to evaluate KNN performance. Bold and underlined numbers highlight the best and second best performance}
\resizebox{0.9\linewidth}{!}{
		\renewcommand{\arraystretch}{1.1}
		\begin{tabular}{ccp{2.0cm}<{\centering}c|p{ 1.6cm}<{\centering}p{ 1.6cm}<{\centering}|p{ 1.6cm}<{\centering}p{ 1.6cm}<{\centering}}

			\hline

			\multirow{2}{*}{\# } & \multirow{2}{*}{\makecell{Feature \\Aggragtor}} & \multirow{2}{*}{\makecell{Feature \\Extraction}} &\multirow{2}{*}{\makecell{Pretrain \\Method}} & \multicolumn{2}{c|}{NSCLC}           & \multicolumn{2}{c}{BRCA }            \\ 
			&   &    &     & Acc.  & AUC   & Acc.  & AUC   \\ \hline
			\multicolumn{7}{l}{\makecell[l]{\cellcolor{mygray}{Weakly supervised classification}}}    \\ \hline
			
            1& \multirow{2}{*}{MIL~\cite{lu2021data}} &  patch-level &  DINO & 0.780$_{\pm 0.126}$ & 0.864$_{\pm 0.089}$ & 0.822$_{\pm 0.047}$   & 0.783$_{\pm 0.056}$    \\
            2 & & region-level &  SLPD    & 0.856$_{\pm 0.025}$ & 0.926$_{\pm 0.017}$ &\textbf{0.879}$_{\pm 0.035}$ & 0.863$_{\pm 0.076}$    \\
            \hline
            
            3& \multirow{3}{*}{DS-MIL~\cite{li2021dual}} &  patch-level &  DINO & 0.825$_{\pm 0.054}$ & 0.905$_{\pm 0.059}$ & 0.847$_{\pm 0.032}$ & 0.848$_{\pm 0.075}$   \\
            4& & region-level &  DINO    & {0.841}$_{\pm 0.036}$ & {0.917}$_{\pm 0.035}$ &0.854$_{\pm 0.032}$& 0.848$_{\pm 0.075}$   \\  
            5& & region-level &  SLPD    & \underline{0.858}$_{\pm 0.040}$ & \underline{0.938}$_{\pm 0.026}$ &0.854$_{\pm 0.039}$& 0.876$_{\pm 0.050}$   \\
            
            \hline
            6 & \multirow{4}{*}{ViT$_{\text{WSI}}$-4096~\cite{chen2022scaling}} & region-level &  DINO & 0.843$_{\pm 0.044}$ & 0.926$_{\pm 0.032}$ &0.849$_{\pm 0.037}$& 0.854$_{\pm 0.069}$   \\
            7 & & region-level &  DINO$+\mathcal{L}_{\text{intra}}$       &  0.850$_{\pm 0.042}$ & 0.931$_{\pm 0.041}$ &  0.866$_{\pm 0.030}$ &   \underline{0.881}$_{\pm 0.069}$ \\
            8& & region-level &  DINO$+\mathcal{L}_{\text{inter}}$       &  0.850$_{\pm 0.043}$ & \underline{0.938}$_{\pm 0.028}$ &0.860$_{\pm 0.030}$& 0.874$_{\pm 0.059}$ \\
            9& & region-level &  SLPD       &  \textbf{0.864}$_{\pm 0.042}$ & \textbf{0.939}$_{\pm 0.022}$ &\underline{0.869}$_{\pm 0.039}$ & \textbf{0.886}$_{\pm 0.057}$ \\ \hline
			\multicolumn{7}{l}{\makecell[l]{\cellcolor{mygray}K-nearest neighbors (KNN) evaluation} }   \\ 
            \hline
            10 & \multirow{4}{*}{Mean} & region-level &  DINO & 0.770$_{\pm 0.031}$ & 0.840$_{\pm 0.038}$ & 0.837$_{\pm 0.014}$ & 0.724$_{\pm 0.055}$  \\
            11 & & region-level &  DINO$+\mathcal{L}_{\text{intra}}$  &  0.776$_{\pm 0.039}$ & 0.850$_{\pm 0.023}$ &0.841$_{\pm 0.012}$& 0.731$_{\pm 0.064}$ \\
            12& & region-level &  DINO$+\mathcal{L}_{\text{inter}}$  & \underline{0.782}$_{\pm 0.027}$  & \underline{0.854}$_{\pm 0.025}$ &\underline{0.845}$_{\pm 0.014}$ & \underline{0.738}$_{\pm 0.080}$  \\
            13& & region-level &  SLPD  &  \textbf{0.792}$_{\pm 0.035}$  & \textbf{0.863}$_{\pm 0.024}$  &\textbf{0.849}$_{\pm 0.014}$   & \textbf{0.751}$_{\pm 0.079}$ \\
            
            \hline
	\end{tabular}}
	\label{tab:main}
\end{table}
\subsection{Weakly-Supervised Classification}
We conduct experiments on two slide-level classification tasks, NSCLC subtyping and BRCA subtyping, and report the results in Tab.~\ref{tab:main}. 
The region-level embeddings generated by SLPD outperform the patch-level embeddings across two aggregators~\footnote{The feature extraction of the patch-level is impracticable for the ViT-based model due to its quadratic complexity in memory usage.} and two tasks (\#1$\sim 5$). This illustrates that learning representations with broader image contexts is more suitable for WSI analysis. Compared with the strong baseline, $i.e.$, the two-stage pre-training method proposed by HIPT (\#6), SLPD achieves performance increases of 1.3\% and 3.2\% AUC on NSCLC and BRCA (\#9). Nontrivial performance improvements are also observed under KNN evaluation (\#10 vs.\#13): $+2.3\%$ and $+3.1\%$ AUC on NSCLC and BRCA. The superior performance of SLPD demonstrates that learning representations with slide-level semantic structure appropriately can significantly narrow the gap between pre-training and downstream slide-level tasks. Moreover, intra-slide and inter-slide distillation show consistent performance over the baseline, corroborating the effectiveness of these critical components of SLPD. 

\begin{table}[t]
	\centering
		\caption{Ablation studies of SLPD. ViT$_{\text{WSI}}$-4096 is the aggregator with region-level embeddings.}
\resizebox{0.9\linewidth}{!}{
		\renewcommand{\arraystretch}{1.1}
		\begin{tabular}{clp{2.0cm}<{\centering}|p{ 1.6cm}<{\centering}p{ 1.6cm}<{\centering}|p{ 1.6cm}<{\centering}p{ 1.6cm}<{\centering}}

			\hline

			\multirow{2}{*}{\# } & \multirow{2}{*}{Ablation} & \multirow{2}{*}{Method} & \multicolumn{2}{c|}{NSCLC}           & \multicolumn{2}{c}{BRCA }            \\ 
			&   &   & Acc.  & AUC   & Acc.  & AUC   \\ \hline
			
            1& \multirow{2}{*}{\makecell[l]{\cellcolor{mygray}{Different cluster-} \\ \cellcolor{mygray}{ing methods}}} &  global  & 0.848$_{\pm 0.045}$ & 0.925$_{\pm 0.033}$ & 0.842$_{\pm 0.048}$   & 0.863$_{\pm 0.060}$    \\
            2 & & slide-level &  0.850$_{\pm 0.042}$ & 0.931$_{\pm 0.041}$ &  0.866$_{\pm 0.030}$ &   0.881$_{\pm 0.069}$ \\
            \hline
            
            3& \multirow{2}{*}{\makecell[l]{\cellcolor{mygray}{Different inter-} \\ \cellcolor{mygray}{slide distillations}}} & region   & 0.828$_{\pm 0.040}$& 0.915$_{\pm 0.025}$ &0.843$_{\pm 0.024}$ &0.849$_{\pm 0.067}$\\
            4& & prototype &  0.850$_{\pm 0.043}$ & 0.938$_{\pm 0.028}$ &0.860$_{\pm 0.030}$& 0.874$_{\pm 0.059}$  \\
            
            \hline
            5 & \multirow{3}{*}{\makecell[l]{\cellcolor{mygray}{Number of} \\ \cellcolor{mygray}{prototypes}}} & $M=2$ & 0.859$_{\pm 0.036}$ &0.936$_{\pm 0.021}$ & 0.869$_{\pm 0.039}$ & 0.886$_{\pm 0.057}$  \\
            6 &  & $M=3$   & 0.864$_{\pm 0.035}$ & 0.938$_{\pm 0.022}$ & 0.861$_{\pm 0.056}$ &  0.878$_{\pm 0.069}$\\
            7& & $M=4$  &  0.864$_{\pm 0.042}$ & 0.939$_{\pm 0.022}$ & 0.860$_{\pm 0.031}$ &  0.872$_{\pm 0.060}$ \\

            \hline
            8 & \multirow{3}{*}{\makecell[l]{\cellcolor{mygray}{Number of} \\ \cellcolor{mygray}{slide neighbors}}} & $K=1$ &   0.864$_{\pm 0.042}$ & 0.939$_{\pm 0.022}$ & 0.869$_{\pm 0.039}$ & 0.886$_{\pm 0.057}$ \\
            9 & & $K=2$ &0.862$_{\pm 0.039}$ & 0.938$_{\pm 0.029}$& 0.875$_{\pm 0.038}$& 0.889$_{\pm 0.057}$     \\
            10 & & $K=3$ & 0.869$_{\pm 0.034}$ & 0.936$_{\pm 0.024}$ & 0.873$_{\pm 0.051}$& 0.880$_{\pm 0.058}$     \\

            \hline
	\end{tabular}}
	\label{tab:ab}
\end{table}
\subsection{Ablation Study}

\noindent \textbf{Different clustering methods.}
As discussed in Sec.~2.3, we can alternatively use the global clustering to obtain prototypes and then optimize the network with a similar distillation objective as Eq.~\ref{eq:intra}. For a fair comparison, the total number of prototypes of the two clustering methods is approximately the same.
Tab.~\ref{tab:ab}(\#1,2) reports the comparative results, where the slide-level clustering surpasses the global clustering by 0.6\% and 1.8\% of AUC on NSCLC and BRCA, which verifies the effectiveness of the former. The inferior performance of the global clustering is due to the visual bias underlying the whole dataset.

\noindent \textbf{Different inter-slide distillations.}
The proposed inter-slide distillation is semantic structure-aware at the slide-level, since we build the correspondence between the region embedding and the matched prototype (\#4 in Tab.~\ref{tab:ab}). To verify the necessity of this distillation method, we turn to another design where the inter-slide correspondence is explored through two nearest region embeddings across slides (\#3 in Tab.~\ref{tab:ab}). As can be seen, the region-level correspondences lead to inferior performances, even worse than the baseline (\#5 in Tab.~\ref{tab:main}), because the learning process is not guided by the slide-level information.

\noindent \textbf{Number of prototypes.}
As shown in Tab.~\ref{tab:ab}(\#5$\sim$7), the performance of SLPD is relatively robust to the number of prototypes on NSCLC, but is somewhat affected by it on BRCA. One possible reason is that the heterogeneity of invasive breast carcinoma is low~\cite{ohlschlegel2011her2}, and thus the excessive number of prototypes cannot obtain medically meaningful clustering results. Empirically, we set $M=4$ on NSCLC and $M=2$ on BRCA as the default configuration. We suggest the optimal number of prototypes should refer to clinical practice, by considering tissue types, cell morphology, gene expression and other factors.

\noindent \textbf{Number of slide neighbors.}
As demonstrated in Tab.~\ref{tab:ab}(\#5$\sim$7), the performance of SLPD is robust to the number of slide neighbors. Considering that more slide neighbors require more computation resources, we set $K=1$ as the default configuration. For more results, please refer to the Supplementary.

\section{Conclusion}
This paper reflects on slide-level representation learning from a novel perspective by considering the intra- and inter-slide semantic structures.  This leads to the proposed Slide-Level Prototypical Distillation (SLPD), a new self-supervised learning approach achieving the more comprehensive understanding of WSIs.
SLPD leverages the slide-level clustering to characterize semantic structures of slides. By representing slides as prototypes,  the mutual-region/slide relations are further established and learned with the proposed intra- and inter-slide distillation. Extensive experiments have been conducted on multiple WSI benchmarks and SLPD achieves state-of-the-art results. Though SLPD is distillation-based, we plan to apply our idea to other pre-training methods in the future, $e.g.$, contrastive learning~\cite{simclr,mocov2}.

\noindent \textbf{Acknowledgement.} This work was supported in part by NSFC 62171282, Shanghai Municipal Science and Technology Major Project (2021SHZDZX0102), 111 project BP0719010, STCSM 22DZ2229005, and SJTU Science and Technology Innovation Special Fund YG2022QN037.
%
%
%
\bibliographystyle{splncs04}
\bibliography{reference}




\end{document}



\section{Supplementary: Visualization Results}
\begin{figure}[h]
\centering
\includegraphics[scale=0.23]{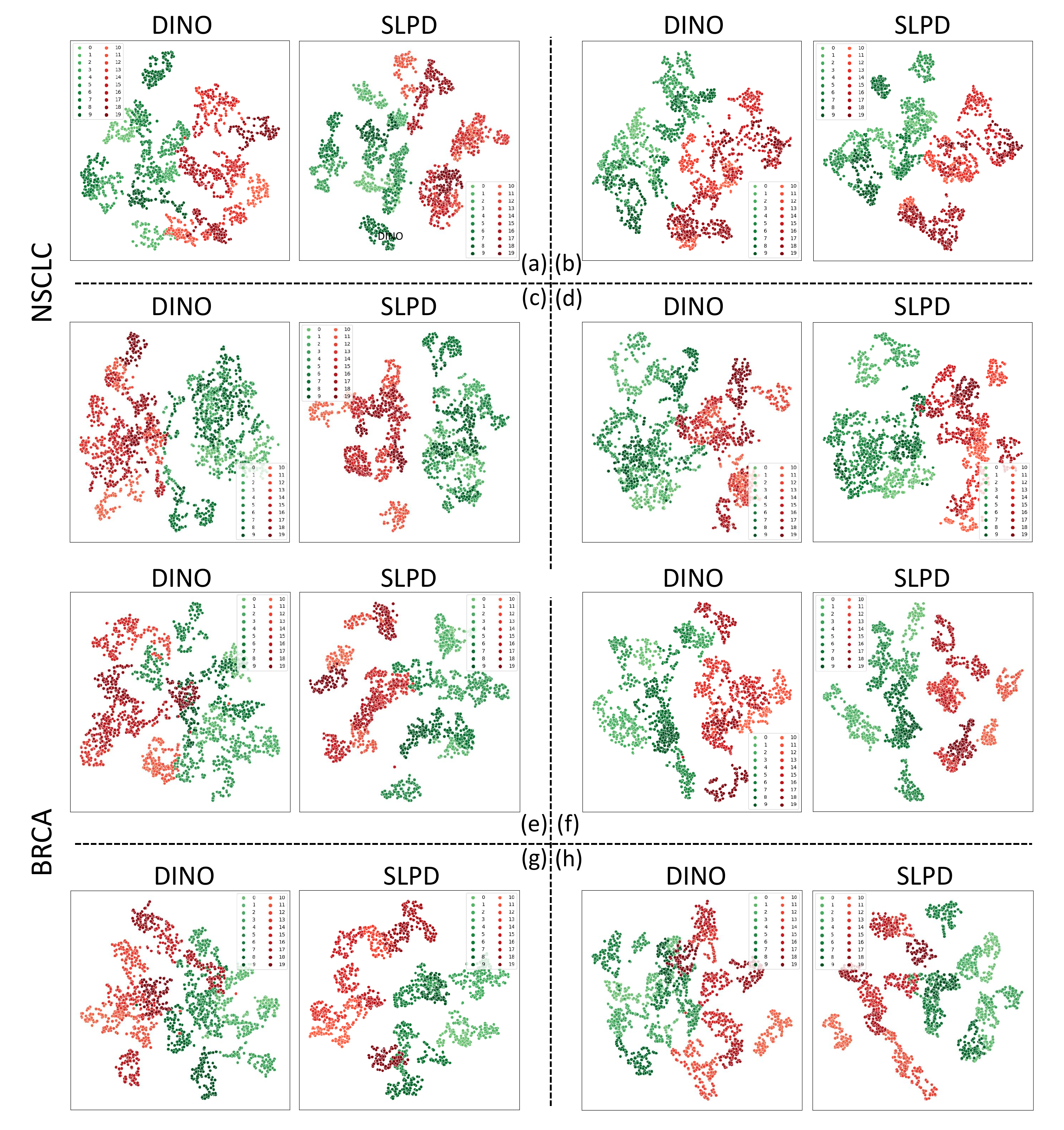}
\caption{For each subfigure, we randomly sample 10 slides from each category of the dataset. Then the region-level representations within each slide are extracted with DINO and SLPD, respectively. We use t-SNE to visualize the representations for NSCLC (a-d) and BRCA (e-h). Two categories are distinguished by red and green. SLPD forms a more separated feature space of different categories and has the more compact cluster at the region-level of each slide. }
\label{fig:TSNE}
\end{figure}